\pdfoutput=1

\documentclass[11pt]{article}

\usepackage[]{ACL2023}

\usepackage{times}
\usepackage{latexsym}

\usepackage[T1]{fontenc}

\usepackage[utf8]{inputenc}

\usepackage{microtype}
\usepackage{inconsolata}
\usepackage{algorithm}
\usepackage{algorithmic}
\usepackage{listings}
\usepackage{newfloat}
\usepackage{listings}
\usepackage{titlesec}
\usepackage{verbatim}
\usepackage{multirow}
\usepackage{graphicx}
\usepackage{booktabs}
\usepackage{amssymb}
\usepackage{amsmath}
\usepackage{bm}
\usepackage{color, soul}
\usepackage[capitalise]{cleveref}

\usepackage[algo2e]{algorithm2e} 

\definecolor{wingreen}{rgb}{0,0.45,0.24}
\definecolor{losered}{rgb}{1.0,0.1,0.24}

\newcommand{\methodname}[1]{FewDocAE}
\newcommand{\basea}[1]{base}
\newcommand{\baseb}[1]{base}
\newcommand{\data}[1]{data}

\title{Few-Shot Document-Level Event Argument Extraction}

\author{Xianjun Yang \qquad Yujie Lu \qquad Linda Petzold  \\
\texttt{\{xianjunyang,petzold\}@ucsb.edu}\\ 
 \texttt{ yujielu10@gmail.com }\\
 Department of Computer Science \\
University of California, Santa Barbara }

\begin{document}
\maketitle

\begin{abstract}
Event argument extraction (EAE) has been well studied at the sentence level but under-explored at the document level. In this paper, we study to capture event arguments that actually spread across sentences in documents. Prior works usually assume full access to rich document supervision, ignoring the fact that the available argument annotation is usually limited.
To fill this gap, we present \textbf{\methodname~}, a \textbf{Few}-Shot \textbf{Doc}ument-Level Event \textbf{A}rgument \textbf{E}xtraction benchmark, based on the existing document-level event extraction dataset. 
We first define the new problem and reconstruct the corpus by a novel $N$-Way-$D$-Doc sampling instead of the traditional $N$-Way-$K$-Shot strategy. Then we adjust the current document-level neural models into the few-shot setting to provide baseline results under in- and cross-domain settings. Since the argument extraction depends on the context from multiple sentences and the learning process is limited to very few examples, we find this novel task to be very challenging with substantively low performance. 
Considering \methodname~ is closely related to practical use under low-resource regimes, we hope this benchmark encourages more research in this direction. Our data and codes will be available online\footnote{\url{https://github.com/Xianjun-Yang/FewDocAE}}.
\end{abstract}

\section{Introduction}
Event argument extraction (EAE), a sub-task of event extraction, is a fundamental task for many downstream NLP applications in the IE community. For example, events and arguments play an important role in the knowledge base population from unstructured data \cite{ge2018eventwiki, li2021document}. And the real world public affairs management relies on recognizing the event arguments from daily news and social media \cite{yuan2018open, ritter2012open}. 
Although tremendous progress has been made under the supervised setting, current neural models typically rely on large-scale human-annotated data, which is not reliable considering the huge amounts of novel events and arguments emerging in many fields every day. Few-shot learning (FSL) \cite{fei2006one} is proposed to tackle such limitations to make machine learning models more applicable given limited annotated examples and has been used a lot in the IE area \cite{han2018fewrel, ding2021few, lai2021learning}.

\begin{figure*}
\centering
    \includegraphics[width=1.0\textwidth]{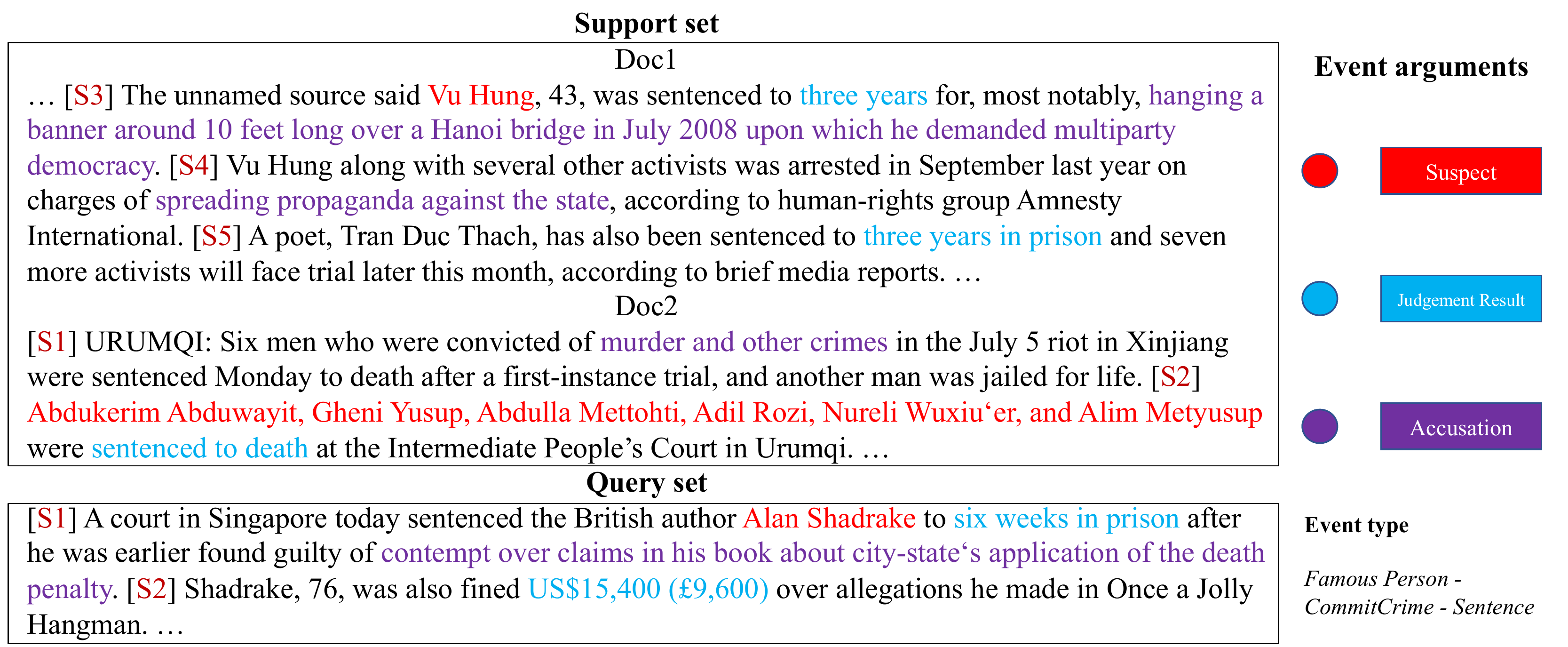} 
    \caption{\label{fig:task} An example of a $3$-Way-$2$-Doc episode consisting of a support and query set. Given a support set of $2$ documents with $3$ argument types, the goal is to extract all event arguments in the query document. During testing, the argument types are disjoint from type set in the training set. But the documents still share the same main event type. }
\end{figure*}

Previous research \cite{yang2019exploring, tong2020improving} mainly focuses on sentence-level event extraction, such as the popular ACE2005 \cite{doddington-etal-2004-automatic} dataset. In recent years, researchers have begun to realize that the complete event and arguments actually spread in a full document or paragraph \cite{li2021document, ebner2020multi, juanzi2022}. And the focus starts to turn into document-level event extraction motivated by the newly proposed large-scale document-level corpora, namely the WikiEvents \cite{li2021document}, RAMS \cite{ebner2020multi} and the recent largest corpus DocEE \cite{juanzi2022}. Following these datasets, many novel methods for solving such new challenges brought by the longer context have also been investigated and witness significant progress \cite{ du2020document, li2021document, xu2022two}. 

However, the traditional supervised learning methods heavily rely on large-scale annotated training data, but we are witnessing new events every day due to the rapid emergence of new affairs. Thus it is not durable to greedily make large collections of newly appeared events for real-life applications. 
Therefore, more attention has been paid to few-shot event extraction. But the current research has only considered the few-shot EAE on a single sentence \cite{lai2021learning, lai2020extensively, deng2020meta}, ignoring the big gap between realistic scenarios. Therefore, we aim to pave a new way for few-shot EAE at the document level towards urgent data scarcity problems on the complete documents. 

The recently released large-scale document-level EE datasets include RAMS \cite{ebner2020multi} and DocEE \cite{tong2022docee}, and their statistics is shown in Table \ref{tab:general-statistics}. RAMS and DocEE contain $139$ and $59$ event types, $65$ and $158$ arguments types, with a total collection of $9,124$ and $21,450$ documents, respectively. While other datasets such as WikiEvents \cite{li2021document} and MUC-$4$ \cite{grishman-sundheim-1996-message} only contain extremely limited types and documents, thus not suitable for our settings. Therefore, we formulate our \methodname~ based on the largest DocEE dataset. Different from FSL for single sentences by traditional $N$-way-$K$-Shot sampling, a novel $N$-Way-$D$-Doc sampling strategy is proposed for our document-level task, as can be seen from the example in Fig. \ref{fig:task}. Besides, previous few-learning problems often fall into the pitfall of robust and fair evaluation, and we follow the FLEX \cite{bragg2021flex} (Few-shot Language Evaluation across(X) many transfer types Principles) to design our settings to avoid such weaknesses. Moreover, prototypical networks (ProtoNet) \cite{snell2017prototypical} have been proven to be very powerful for solving few-shot problems by representing each category as a prototype in both Vision \cite{pan2019transferrable, dong2018few} and NLP \cite{sun2019hierarchical, gao2019hybrid} domains. We combine ProtoNet and a pre-trained language encoder for establishing baselines on our new task and provide a comprehensive analysis. 

The key contributions of this work include:
\begin{itemize}
\item We are the first to introduce few-shot document-level event argument extraction, greatly extending supervised document-level EAE to few-shot scenarios.
\item We reconstruct a realistic \methodname~ dataset along with a new few-shot sampling algorithm, $N$-Way-$D$-Doc sampling.
\item We conduct comprehensive experiments to provide benchmark results and find the tasks are extremely challenging and worth further investigation.
\end{itemize}

\section{Related Work}

\subsection{Document-Level Information Extraction}
In the IE community, previous work mainly focuses on sentence-level tasks. For example, the commonly-used relation extraction benchmark TACRED \cite{zhang2017position}, the ACE05 \footnote{https://catalog.ldc.upenn.edu/LDC2006T06} and KBP$2017$ \footnote{https://tac.nist.gov/2017/KBP/} event extraction datasets, the CoNLL-$2003$ \cite{sang2003introduction} and OntoNotes $5.0$ \cite{pradhan2013towards} named entity recognition corpora, all focus on single sentence-level semantics. Since information extraction often involves document-level reasoning, recently there have been great efforts to contribute document-level benchmarks. For instance, the DocRED \cite{yao2019docred} is the largest dataset to extend relation extraction to the document level. The recent RAMS \cite{ebner2020multi} and DocEE \cite{tong2022docee} corpora focus on multi-sentence event extraction. 
Although tremendous progress has been made in the information extraction area on sentence-level tasks, the emerging document-level datasets raise new challenges. The difficulty mainly comes from the long context semantic representation brought by multiple sentences and the extremely unbalanced label distribution.

\subsection{Document-Level Event Argument Extraction}
Event extraction can be classified into trigger-word and no-trigger word based extraction, including event detection and event arguments extraction. 
Many approaches and datasets \cite{petroni2018extensible, hurriyetoglu-etal-2021-multilingual, giorgi-etal-2021-discovering, zavarella-etal-2022-tracking} across diverse domains have been proposed for document-level argument extraction to go beyond single-sentence inference. For example, \cite{ebner2020multi} build RAMS to include cross-sentence argument annotations but still limits the arguments around the event in a $5$-sentence window. An end-to-end generative transformer \cite{du2021template} regards argument extraction as a template-filling task. Besides, \cite{li2021document} formulates the task as conditional generation following event templates, and contributes to the WIKIEVENTS dataset, which consists of only $246$ documents with less than one-fourth of annotated cross-sentence arguments. Very recently, \cite{du2022dynamic} introduce a new global neural generation-based framework by constructing a document memory store to record the contextual event information for improving capability. The largest document-level event extraction dataset is the DocEE \cite{tong2022docee}, which consists of $27,000+$ events, $180,000+$ arguments over $27,485$ Wikipedia articles. In this paper, we use the DocEE as base set for our task.

\subsection{Few-Shot Argument Extraction}
Few-shot learning for information extraction is proposed to tackle such circumstances when only limited instances are annotated. There have been growing interests under few-shot settings for named entity recognition \cite{ding2021few, das2022container}, and relation extraction \cite{han2018fewrel, DBLP:conf/naacl/Popovic022} under single-sentence and document-level scenarios. There has also been research \cite{deng2020meta, lai2021learning, DBLP:journals/corr/abs-2010-11325, lai2020extensively} for few-shot event extraction within single-sentence. However, to the best of our knowledge, there is no research about document-level few-shot event argument extraction until the submission date.

To fill this gap, this work focuses on few-shot learning for document-level argument extraction. Instead of building a new dataset from scratch, we aim at leveraging the existing supervised dataset for reconstructing the instances by a novel $N$-Way-$D$-Doc sampling strategy, inspired by similar work \cite{ sabo2021revisiting, DBLP:conf/naacl/Popovic022}.

\section{Task Formulation}

\subsection{Argument Extraction Definition}
The event argument extraction usually depends on the first-stage detected events, but we assume gold event labels to reduce the task complexity in our work. This is opposite to joint extraction where the task is to jointly extract all events and their associated arguments all at once \cite{sha2018jointly, yang2016joint}.
Since the DocEE dataset \cite{tong2022docee} follows the main event extraction \cite{DBLP:conf/iconference/HamborgLSHG18} setting where no trigger words exist and the article title $t$ and the article $a$ itself together determine the event type, we follow their setting and assume the event type $e$ is given, then aim at extracting all related arguments with types $R_e$.

Formally, given a document $D=\{w_1, ..., w_{|D|}\}$ and its corresponding event type $e$, where $|D|$ is the total number of words, the event argument extraction aims to detect the boundaries and types for all possible continuous spans $\{w_{start}, w_{end}\}$ in the document $D$ according to event argument types $R_e$. 

\begin{table*} 
\centering
\begin{tabular}{lccccccc}
\hline
\textbf{Data Set} & \textbf{\# Docs.} & \textbf{\# ET.} & \textbf{\# AT.} & \textbf{\# Tok/Doc} & \textbf{\# Sents/Doc} & \textbf{\# ArgInst.} & \textbf{\#ArgScat.} \\
\hline
DocEE & $21,450$ & $59$ & $358$ & $678$ & $30.71$ & $109,395$ & $10.2$ \\
RAMS & $9,124$ & $139$ & $65$ & $105$ & $3.79$ & $21,237$ & $4.8$ \\
\hline
\end{tabular}
\caption{\label{tab:general-statistics} A comparison of DocEE \cite{juanzi2022} and RAMS \cite{ebner2020multi}. Docs.: document, ET.: event type, AT.: event argument type, Tok/Doc: tokens per document, Sents/Doc.: sentences per document, ArgInst.: event arguments, ArgScat.: the
number of sentences in which event arguments of the same event are scattered.}
\end{table*}

\begin{table} 
\centering
\scalebox{0.74}{
\begin{tabular}{c|cc|cc|cc}
\hline
\multirow{2}{*}{\textbf{Types}} & 
\multicolumn{2}{c|}{ Train } & \multicolumn{2}{c|}{Dev} & \multicolumn{2}{c}{ Test} \\
 & \#ET. & \#AT. & \#ET. & \#AT. & \#ET. & \#AT. \\
 \hline
{In domain (small)} & $30$ & $193$ & $14$ & $87$ & $15$ & $68$ \\
{In domain (base)} & $49$ & $303$ & $10$ & $51$ & $10$ &  $50$ \\
{Cross domain}  & $49$ & $302$ & $10$ & $53$ & $10$ & $54$ \\
\hline
\end{tabular}
}
\caption{\label{split stats} The statistics of chosen event and argument types in our three domain split settings. }
\end{table}

\subsection{Document-Level Few-Shot Argument Extraction} \label{task: event}
Following previous work about sentence-level few-shot event detection \cite{deng2020meta},
we define the document-level few-shot argument extraction as the following. Given the event instance $e$, its associated argument types set $R_e$, the support set $S$ and the query set $Q$, the few-shot task $T$ is defined as:
$$S = \{..., R_i^s, ... \}, R_i^s= (D_i^s, \{..., (b_i^s, t_i^s), ...)\}$$
$$Q = \{..., R_i^q, ... \}, R_i^q= (D_i^q, \{..., (b_i^q, t_i^q), ...\})$$
$$T = \{S, Q\}$$
where $(b_i, t_i)$ represents the $i$-th event argument boundaries and type in document $D_i$ in the support $s$ and query $q$ set. $R_i^{s/q}$ is the set of all the annotated arguments in $D_i$ and $S/Q$ is the combinations of $R_i^{s/q}$ from different documents.
Following the episode training for few-shot learning, a task $T$ is one episode aiming to predict all the instances in $Q$ given $S$. The few-shot learning is usually formalized as an $N$-Way-$K$-Shot problem, which means that there are $N$ possible argument types and $K$ supporting instances for each argument type for every task $T$. Note that the argument types set $R_\textbf{train}$ and $R_\textbf{test}$ are disjoint.

\begin{table} 
\centering
\scalebox{0.68}{
\begin{tabular}{c|cc|cc| cc}
\hline
\multirow{2}{*}{\textbf{Avg. Args}} & 
\multicolumn{2}{c|}{\textbf{In domain (small)}} & \multicolumn{2}{c|}{\textbf{In domain (base)}} & \multicolumn{2}{c}{\textbf{Cross domain}} \\
 & micro & macro & micro & macro  & micro & macro \\
\hline
$3$w$1$d & $4.41$ & $4.23$ & $4.40$ & $4.58$ &  $4.47$ & $4.35$\\
$3$w$2$d & $3.20$ & $3.22$ & $3.57$ & $3.11$ &  $3.52$ & $3.03$ \\
$6$w$2$d & $6.56$ & $6.35$ & $7.16$ & $6.28$ &  $6.56$ & $6.35$ \\
\hline

\end{tabular}
}
\caption{\label{nwdd stats} The average number of arguments in different settings in the training episodes. The micro average is calculated on average across all episodes on D-doc. The macro average is calculated on average across all argument types on $D$-doc. }
\end{table}

In practice, given all the support documents in $S$, we want to extract all the arguments for documents in $Q$. Since it is not guaranteed each support instance $D_i^s$ contains only one argument, the $K$ in the traditional $N$-Way-$K$-Shot setting is no longer guaranteed to be an integer. Previous research \cite{yang2020simple} tries to use greedy sampling to guarantee the strict $K$ shots requirements for sentence-level few-shot NER task, but this is not applicable due to the sparse density of arguments in the document as also been observed by \cite{ding2021few}. And they loose the $K$ shots requirement to $K$$\sim$$2K$ shots. However, this is still not realistic under a document-level setting since the arguments spread become even sparser and $K$$\sim$$2K$ shots are still not guaranteed. It is notable that \cite{DBLP:conf/naacl/Popovic022} tackle this problem for few-shot document-level relation extraction by $D$-Doc setting where both $N$ and $K$ are variables between documents and individual episodes. We argue that variable $N$ is not suitable for deploying applications and will make models complicated, so we design a novel $N$-Way-$D$-Doc sampling, where $N$ and $D$ are both fixed, resulting in variable $K$ shots.

\subsection{Domain Split}
To make our task closer to realistic applications, we consider three settings for investigating the difficulty of tasks and model performance.
In the \emph{In-domain} circumstance, we manually choose event and event argument types from the same coarse-grained label sets to ensure that they share the same domain knowledge. To explore the meta-learning ability with varying amounts of training bases, we further set \emph{In domain (small)} and \emph{In domain (base)} with a small and medium numbers of event types contained in the training set.
The authors \cite{tong2022docee} provide \emph{Cross domain} scenario, where the training and test labels are entirely disjoint, sharing no mutual domain information. We follow their splits for our domain adaptation task.

\begin{algorithm}[t]
\caption{N-Way-D-Doc Sampling for Few-Shot Episode Generation}
\label{algo}
\begin{algorithmic}[1]

\REQUIRE ~~\\ 
Dataset $\mathcal{X}$, Label set $\mathcal{Y}$, $N$, $D$\\
\ENSURE ~~\\ 
\STATE $\mathcal{S} \leftarrow \varnothing$;  // Init the support set
\STATE $\text{CountN} \leftarrow 0$; //Init the count of entity types;
\STATE $\text{CountD} \leftarrow 0$; //Init the count of documents;

\WHILE{$\text{CountN} \neq N$ or $ \text{CountD} \neq D $}
\STATE Randomly sample $(\bm{x}, \bm{y}) \in \mathcal{X}, \mathcal{Y}$ ; 
\STATE Compute $\text{CountN}$ and $\text{CountD}$;
\IF{$\text{CountN} > N$ or $\text{CountD} > D$}
\STATE Continue;
\ELSE
\STATE $\mathcal{S} = \mathcal{S} \bigcup (\bm{x}, \bm{y})$ ;
\STATE Update $\text{CountN}, \text{CountD}$ ;
\ENDIF
\ENDWHILE

\end{algorithmic}
\end{algorithm}

\vspace{-0.2cm}

\section{Constructing Few-Shot Episodes}
In this section, we talk about the details regarding the dataset conversion and how we obtain episodes for training. 
For our \methodname~ task, the key components for constructing a realistic few-shot dataset are made of two parts: avoiding data leakage and constructing effective support and query pairs. 
Generally, to avoid leaking new event arguments information in the training phase we replace all other labels of arguments contained in the validation/test sets with $\texttt{O}$. In this way, it is guaranteed that they have disjoint argument types sets, and this is also close to the realistic scenarios. To provide each episode with a support and query set, we greedy sample instances to make sure they satisfy our $N$-Way-$D$-Doc choice. Also, they support and query instances should always come from different documents.

\subsection{Choosing Datasets }
We initially consider the two largest document-level event extraction datasets for our tasks: RAMS \cite{ebner2020multi} is annotated in a $5$-sentence window around each event trigger and contains $9,124$ annotated events from news based on an ontology of $139$ event types and $65$ roles. In addition, DocEE includes $21,450$ document-level events with $109,395$ arguments, making it the largest document-level event extraction dataset. Since we aim at building a document-level benchmark, we finally exclude RAMS for its $5$-sentence limits and narrow argument types. Eventually, we choose DocEE for our ~\methodname~ task from their original release \footnote{https://github.com/tongmeihan1995/DocEE}.
Besides, in order to make sure there are enough examples in each episode for support and query, we exclude all events and argument types that have lower than $2$ annotated examples in the train/validation/test set.

\subsection{Determining Event Arguments Types}\label{arg types}
In the released DocEE corpus, there are $31$ hard news event types and $28$ soft news event types with their corresponding arguments. Their original paper follows no-trigger words design \cite{nguyen2016dataset, zheng2019doc2edag} and assumes one main event per document.
For arguments, there are $358$ event argument types belonging to those $59$ event types. The argument annotation is done on the whole document resulting in some documents could contain up to $40$ sentences and $7k$ words. 

Since different events could still contain the same argument types, we could mask the overlapped arguments in two ways. The first one masks all arguments in the training set with $\texttt{O}$ if they also appear in the val and test sets as used in \cite{ding2021few}, while in the second strategy, we can mask all the arguments in the val and test sets if they are shared by the training set. The intuition is that the former one has the risk that the model is trained with $\texttt{O}$ types, but is forced to predict their true labels which are not $\texttt{O}$ during testing. We believe the latter setting reduces the difficulty by making sure that the new arguments appearing during the prediction stage are not labeled with $\texttt{O}$ during training, and use this strategy for all our experiments.

For the \textbf{In domain (small)} setting, we manually choose $14$ and $15$ event types for constructing the validation and test episodes, while leaving all other $32$ event types for training purposes. This results in disjoint event sets for train/val/test, but they share some domain knowledge. Since the number of event types used for training is only about two times that of testing, we believe this setting is more challenging. The intuition is that for few-shot learning, we aim to get a good feature extractor during training using massive base data so that those learned features could benefit the model by predicting novel instances. We use this setting for exploring the few-shot learning limitation when the training base is small.
For the \textbf{In domain (base)} setting, we use a larger training base with $49$ event types and adopt the remaining $10$ event types for validation and testing. Again, our choice of event splitting guarantees they share in domain event types. In contrast to previous situation, we now aim at investigating the full capability of few-shot learning given enough training base.

For \textbf{Cross domain}, we follow the original event splits in DocEE where the authors choose the natural disasters events as the target domain, including $\texttt{Floods}$, $\texttt{Droughts}$, $\texttt{Earthquakes}$, $\texttt{Insect Disaster}$, $\texttt{Famine}$, $\texttt{Tsunamis}$, $\texttt{Mudslides}$, $\texttt{Hurricanes}$, $\texttt{Fire}$, and $\texttt{Volcano Eruption}$, and leave the remaining $49$ event types as source domains. 

Besides, there are six arguments, including $\texttt{Date}$, $\texttt{Causes}$, $\texttt{Areas affected}$, $\texttt{Location}$, $\texttt{Casualties}$, and $\texttt{Losses}$ that occur frequently in all the splits. To fully leverage the capability of meta-learning during the training phase, we leave those arguments in the training set only and mask them as $\texttt{O}$ in the val and test sets. The full statistics of our domain split result is shown in Table~\ref{split stats}.

\subsection{Sampling Strategy}
The traditional $N$-Way-$K$-Shot sampling for few-shot learning fails when applied to our document-level settings. The reason is that one document with the sparse spread of event arguments could contain one or many arguments. Thus greedy sampling $K$ shots instance can not be guaranteed.
Soft sampling methods like $N$-Way-$K$$\sim$$2K$ Shots in \citep{ding2021few} still do not work for our long documents since $K$$\sim$$2K$ Shots are still hard to be satisfied. Note that \citep{DBLP:conf/naacl/Popovic022} adopts $D$-doc sampling for document-level relation extraction. However, their approach can not guarantee a fixed number of $N$ classes. For example, their $1$-Doc results in $2.18$-Way-$2.36$-Shot instances on average. We argue that variable $N$ is not ideal for our task.

We introduce a new sampling strategy, $N$-Way-$D$-Doc sampling as shown in Algorithm \ref{algo}, to guarantee a fixed number of $N$ classes within $D$ documents. In our approach, we first pick $N$ event argument types, then randomly sample $D$ documents and keep all the arguments within $N$ types in the $D$ documents unchanged while discarding other argument types. This process is not finished until there are exactly $N$ classes within $D$ documents. Besides, few-shot learning is notoriously famous for lacking challenging-yet-realistic testing setups and failing to employ careful experimental design \cite{bragg2021flex}. To make our benchmark more robust, we sample episodes by making sure they evenly come from combinations of different documents, events, and argument types. 

\subsection{Sampling Results}
We sample around $30k$$\sim$$50k$ episodes for the training and around $3k$ episodes for validation and testing sets under $3$-Way-$2$-Doc and $6$-Way-$2$-Doc settings. For the $3$-Way-$1$-Doc setting, since not all documents have at least three unique arguments, we sample about $30k$ for training and $1.5k$ for Val and test.
We also show the argument type characteristics of our various $N$-Way-$D$-Doc sampling results in Table \ref{nwdd stats}. As we can see, the average sampled number of arguments for both settings is close. For the $3$-Way-$1$-Doc settings, there are around $4.47/3 = 1.5$ arguments per type, while only around $1.14$ and $1.10$ arguments per type for other settings, demonstrating the diversity of our splitted domains.

\subsection{N-Way-D-Doc Choice}
Due to the unique nature of few-shot document-level extraction, we only test limited combinations of N-way and D-doc for two reasons. First, there are only limited documents that simultaneously contain the same N argument types, thus we can not find enough samples when extending to more than 6-way. Second, we only keep 1/2-doc for the two reasons: on the one hand, extending to more documents requires doing sequence labelling on much longer documents that consumes much more GPU memory that we struggle to handle even by a 40G GPU. Similar memory issue has also been reported by \cite{sabo2021revisiting} when handling few-shot learning for relation extraction.
On the other hand, adding more documents will also increase the number of NOTA argument types, which will be detrimental for extracting useful real arguments since the vast majority would be None types. Considering the current settings already result in relatively low performance, we do not aim to further increase the challenge.

\section{Models}
Previous work on document-level EE using BERT\_Seq \cite{du2020document, tong2022docee} demonstrate the success of using a pre-trained BERT model to sequentially label words in the article. And the superior performance of the long document transformer (e.g. Longformer \cite{DBLP:journals/corr/abs-2004-05150}) has also been proven to improve the argument extraction task \cite{tong2022docee}. We thus follow their baseline settings and use BERT or Longformer as document encoders for our task. 
In order to adapt to our few-shot setting, inspired by the successful applications of the prototypical network (ProtoNet) \cite{snell2017prototypical} for meta-learning, we assume there exists one prototypical representation for each argument type. Then we implement the models by extending ProtoNet to language encoder with token-level similarity.

\begin{table*}
\centering
\scalebox{0.66}{
\begin{tabular}{l@{\hspace{0.8\tabcolsep}}|ccc|ccc|ccc|ccc}
\hline
\multirow{2}{*}{\textbf{Model}}
 & \multicolumn{3}{c|}{\textbf{Baseline}} & \multicolumn{3}{c|}{\textbf{ProtoNet-BERT}} & \multicolumn{3}{c|}{\textbf{ProtoNet-LongFormer}} & \multicolumn{3}{c}{\textbf{ProtoNet-MNAV}}   \\
  \cline{2-13}
 &  P  & R  & $F_1$ & P  & R  & $F_1$ & P  & R  & $F_1$ & P  & R  & $F_1$ [\%]\\
\hline
 $3$-Way-$1$-Doc & $0.88$ & $2.50$ & $1.30$ & $3.31 \pm 1.44$ & $15.52 \pm 1.45$ & $5.35 \pm 1.87$ & $4.83 \pm 0.39$ & $15.67 \pm 2.66$ & $\textbf{7.39} \pm \textbf{0.65}$ &4.46 & 11.42 & 6.42\\
 $3$-Way-$2$-Doc & $1.22$ & $12.99$ & $2.23$ & $4.77 \pm 0.54$ & $15.2 \pm 0.56$ & $7.26 \pm 0.49$ &  $5.88 \pm 1.58$ & $14.65 \pm 0.39$ & $\textbf{8.34} \pm \textbf{1.59}$ & 5.12 & 13.25 & 7.38 \\
 $6$-Way-$2$-Doc & $0.64$ & $7.58$ & $1.19$ & $4.42 \pm 0.45$ & $11.59 \pm 0.46$ & $6.37 \pm 0.24$ & $6.17 \pm 0.54$ & $14.93 \pm 0.41$ & $\textbf{8.73} \pm \textbf{1.03}$ & 5.73 & 15.08 & 8.30 \\
\hline
\end{tabular}
}
\caption{\label{in-domain small} \emph{In domain (small)} results for \methodname~ argument extraction task under three settings.
}
\end{table*}

\subsection{Document Encoder}
We adopt Longformer (led-base-$16384$\footnote{https://huggingface.co/allenai/led-base-16384}) and BERT (bert-base-uncased\footnote{https://huggingface.co/bert-base-uncased}) as our encoder for all the experiments. In order to handle long sequences, we split all inputs with a chunk length of $1024$ and $512$, respectively.

Formally, suppose the document $D = \{w_1, ..., w_{|D|}\}$ where $w_i$ represents the $i\-th$ token and $|D|$ is the maximum length. By feeding the tokens into the document encoder, we get the contextualized token representation:
$$[h_1, ..., h_{|D|} ] = Encoder([w_1, ..., w_{|D|}])$$

\subsection{Prototypical Networks (ProtoNet) }
The ProtoNet approach \cite{snell2017prototypical} assumes there exists one prototypical representation for each argument class and learns a metric space where categorization is performed by labeling each query term with the value calculated from the distance between prototype representations that are closest to it. In practice, we use the average representation of all tokens in each argument to represent the contextualized representation of that argument type.
Formally, for support set $S_{r_i}$ containing all the arguments of type $r_i$, following previous step $h_t$ is the representation of token $t$. By calculating all the prototypes,
$$r_i = \frac{ \sum_{t\in S_{r_i}} {h_t}}{|S_{r_i}|}$$
we get the argument representation set $R = \{r_1, ..., r_N, r_{N+1}\}$ where $r_i$ represent the $i\-$th argument prototype and $r_{N+1}$ represents the $\texttt{O}$ type.
Then for the query token $h_q$, the target label is assigned as $$label = \mathop{\arg\min}_{i} (d(r_i, h_q))$$.
We use the L$2$ distance as the distance metric. 

Besides, we also add a \textbf{Baseline} model based on ProtoNet-LongFormer without any finetuning. The reason is that training on many episodes is very costly as will be shown below. And we are interested on the original performance and how much benefits the finetuning could bring in.
\subsection{Nearest Neighbor Tagger (NNShot)}
NNShot \cite{yang2020simple} is a simple but strong system based on token-level nearest neighbor classification for the few-shot sequence tagger task. It first obtain contextual representations for all tokens in their respective documents. Then it assigns the token $q$ a tag $r_i$ corresponding to the most similar token in the support set:
$$label = \mathop{\arg\min}_{i} (\arg\min_{i'\in{S_{r_i}}} (d(r_i', h_q)))$$
where $S_{r_i}$ represents the set of support tokens with $r_i$ tags.

However, simply extending the sentence-level NNShot to a long document is not plausible. For example, for two documents both with $4,096$ tokens and each token has an embedding dimension of $768$, the token-level similarity computation requires more than $50$GB GPU memory. To make  token similarity calculation of long inputs possible, we use another linear layer to reduce the $768$ \-dimensional representation into $32$ \-dimension and use the L$1$ distance as the distance metric for computational efficiency. 

\subsection{ProtoNet-MNAV}
Here we adjust the Multiple NOTA(None-Of-the-Above) Vectors(MNAV) proposed by \cite{sabo2021revisiting} for few-shot relation extraction to our \methodname~ since they both face the same issue that the majority labels belong to NOTA. Instead of initializing the NONA vectors by randomly as by \cite{sabo2021revisiting} or from sampled support sets as in \cite{DBLP:conf/naacl/Popovic022} and then gradually update them, we adopt a K-means MNAV strategy. Specifically, we perform a K-means clustering for all NOTA representations where K is set to be a hyperparameter. Then for ProtoNet-based models, we still determine the label type by calculating their token similarity. And we attribute all token nearest to the K NOTA vectors as NOTA type. This way we ideally reduce the risk of only representing many NOTA token by one vector as also pointed by \cite{allen2019infinite}, since there might be multiple NOTA Prototypes. 
We always use LED as the encoder for ProtoNet-MNAV method.

\section{Experiments and Results}\label{analysis}
We show the experimental results and analysis in this section. More experimental details and configurations can be found in Appendix \ref{experiments}. In general, one experiment costs around 15 hours and in total there are more than 100 runs, which also limits more advanced models that we can choose.
\begin{table*}
\centering
\scalebox{0.66}{
\begin{tabular}{l@{\hspace{0.8\tabcolsep}}|ccc|ccc|ccc|ccc}
\hline
\multirow{2}{*}{\textbf{Model}}
 & \multicolumn{3}{c|}{\textbf{Baseline}} & \multicolumn{3}{c|}{\textbf{ProtoNet-BERT}} & \multicolumn{3}{c|}{\textbf{ProtoNet-LongFormer}} & \multicolumn{3}{c}{\textbf{ProtoNet-MNAV}} \\
  \cline{2-13}
 &  P  & R  & $F_1$ & P  & R  & $F_1$ & P  & R  & $F_1$ & P & R & $F_1$ [\%] \\
\hline
 $3$-Way-$1$-Doc & $0.91$ & $5.95$ & $1.58$ & $2.90 \pm 0.52$ & $14.12 \pm 1.11$ & $4.80 \pm 0.78$ &  $6.55 \pm 0.52$ & $19.13 \pm 0.51$ & $\textbf{9.76} \pm \textbf{0.56}$ & 5.08 &  14.49 &  7.52 \\
 $3$-Way-$2$-Doc & $1.91$ & $7.92$ & $3.08$ & $4.87 \pm 0.57$ & $23.05 \pm 0.18$ & $8.03 \pm 0.76$ &  $7.92 \pm 0.58$ & $20.89 \pm 0.93$ & $\textbf{11.48} \pm \textbf{0.89}$ & 7.13 &  20.35 &  10.56 \\
 $6$-Way-$2$-Doc & $3.19$ & $14.77$ & $5.25$ & $2.25 \pm 0.85$ & $14.32 \pm 1.76$ & $4.25 \pm 1.61$ & $8.21 \pm 0.49$ & $22.76 \pm 0.66$ & $\textbf{12.07} \pm \textbf{0.48}$ & 8.54 &  17.68 &  11.52 \\
\hline
\end{tabular}
}
\caption{\label{in-domain base} \emph{In domain (base)} results for \methodname~ argument extraction task under three settings.
}
\end{table*}

\begin{table*}
\centering
\scalebox{0.66}{
\begin{tabular}{l@{\hspace{0.8\tabcolsep}}|ccc|ccc|ccc|ccc}
\hline
\multirow{2}{*}{\textbf{Model}}
 & \multicolumn{3}{c|}{\textbf{Baseline}} & \multicolumn{3}{c|}{\textbf{ProtoNet-BERT}} & \multicolumn{3}{c|}{\textbf{ProtoNet-LongFormer}} & \multicolumn{3}{c}{\textbf{ProtoNet-MNAV}}\\
  \cline{2-13}
 &  P  & R  & $F_1$ & P  & R  & $F_1$ & P  & R  & $F_1$ & P  & R  & $F_1$ [\%] \\
\hline
 $3$-Way-$1$-Doc & $1.12$ & $8.00$ & $1.96$ & $1.94\pm 0.01$ & $12.45 \pm 0.02$ & $ 3.36 \pm 0.01$ & $5.65 \pm 0.78$ & $19.38 \pm 1.52$ & $\textbf{8.70} \pm \textbf{0.75}$ & 5.34 & 17.95 & 8.23 \\
 $3$-Way-$2$-Doc & $1.03$ & $11.22$ & $1.90$ & $2.81 \pm 0.25$ & $22.25 \pm 0.34 $ & $5.34 \pm 0.35$ & $7.09 \pm 0.87$ & $20.52 \pm 0.90$ & $\textbf{10.51} \pm \textbf{0.95}$ & 6.18 & 19.44 & 9.38 \\
 $6$-Way-$2$-Doc & $0.73$ & $7.86$ & $1.34$ & $1.67 \pm 0.47$  & $17.03 \pm 6.37$ & $ 3.05 \pm 0.87$ & $5.91 \pm 0.30$ & $17.58 \pm 0.20$ & $\textbf{8.84} \pm \textbf{0.31}$  & 5.36 & 16.89 & 7.49 \\
\hline
\end{tabular}
}
\caption{\label{cross-domain} \emph{Cross domain} results for \methodname~ argument extraction task under three settings.
}
\end{table*}

\begin{table}
\centering
\scalebox{0.65}{
\begin{tabular}{l@{\hspace{0.8\tabcolsep}}|c|ccc}
\hline
\multirow{3}{*}{\textbf{Settings}}
 & \multicolumn{4}{c}{\textbf{NNShot-LongFormer}}  \\
 \cline{2-5}
 & val & \multicolumn{3}{c}{test}  \\
  \cline{2-5}
 & $F_1$ & Precision  & Recall  & $F_1$ [\%] \\
\hline
 $3$-Way-$1$-Doc & $6.02 \pm 0.89$ & $3.84 \pm 0.67$ & $12.75 \pm 0.78$ & $5.9 \pm 0.93$ \\
 $3$-Way-$2$-Doc & $3.34 \pm 0.43$ & $2.59 \pm 0.40$ & $6.60 \pm 0.43$ & $3.73 \pm 0.46$  \\
 $6$-Way-$2$-Doc & $4.87 \pm 0.68$ & $3.21 \pm 0.66$ & $8.07 \pm 0.67$ & $4.59 \pm 0.86$  \\
\hline
\end{tabular}
}
\caption{\label{cross-domain nnshot} \emph{Cross domain} results for \methodname~ argument detection task under three settings by NNShot model.}
\end{table}

\begin{table} 
\centering
\scalebox{.9}{
\begin{tabular}{c|ccc|ccc}
\hline 
 & \multicolumn{3}{c|}{\textbf{In domain (base)}} & \multicolumn{3}{c}{\textbf{Cross domain}} \\
 & $3$w$1$d & $3$w$2$d & $6$w$2$d & $3$w$1$d & $3$w$2$d & $6$w$2$d \\
\hline
FP & $3.68$ & $2.45$ & $4.86$ & $6.42$ & $3.23$ & $4.40$ \\
FN & $1.99$ & $1.26$ & $1.66$ & $2.60$ & $0.87$ & $1.43$ \\
\hline
\end{tabular}
}
\caption{\label{FN} The FP means a token with true label $\texttt{O}$ is misclassified as a part of an argument, while the FN represents a token within a certain argument is misclassified as $\texttt{O}$. }
\end{table}

\paragraph{In Domain}
For the \emph{In domain} setting where the training and test examples are sampled from the same domain, the performance is exhibited in Table~\ref{in-domain small} and Table~\ref{in-domain base} for the \emph{In domain (small)} and \emph{In domain (base)} results, respectively. As we can see, the baseline without finetuning can not guarantee good performance, and the ProtoNet-Longformer models consistently outperform the ProtoNet-BERT by a large margin under three $N$-Way-$D$-Doc settings. On the one hand, this gap can be explained by the superior encoding ability of long documents by LongFormer. On the other hand, the results convince our motivation by extending to document-level argument extraction as a large portion of arguments can only be extracted across sentences, as also confirmed under supervised condition by \cite{tong2022docee}. Besides, we also observe a significant performance drop when moving from \emph{In domain (base)} to \emph{In domain (small)} under most results, which clearly manifests the benefits of using training base with broader event types. This validates the intuition that more diverse training base can help train a better argument feature extractor. Also, we can see that as we increase the Ways of $N$ and Docs of $D$, the overall results continue increasing, demonstrating instances with more ways and more Docs are easier to be predicted. However, ProtoNet-MNAV does not bring performance gain as expected, possibly due to the more unclear decision boundary as illustrated in Appendix \ref{visualizations}.
In general, the current models remain relatively low results, demonstrating the challenges of our \methodname~ task in document-level. We additionally show some case study in Appendix \ref{case}.

\paragraph{Cross Domain}
The overall arguments extraction results are shown in Table~\ref{cross-domain} for the \emph{Cross domain} setting. Compared with its \emph{In domain (base)} counterpart, the performance degradation is witnessed under both ProtoNet-Longformer and ProtoNet-BERT models in all settings. This is expected since we split the cross domain to avoid the in domain knowledge and such domain adaptation is more challenging.

We also test the performance using NNShot-Longformer in Table \ref{cross-domain nnshot} under \emph{Cross domain}. To be compatible with computation memory limits brought by NNShot token-level similarity calculation, we do not take whole documents and split the documents within chunks of $1024$ tokens. 
For the $1$-Doc setting, an average F1 score of $5.9$\%, and for the $3/6$-Way-$2$-Doc settings, F1 of $3.73$\% and $4.59$\% points are observed using NNShot model. The performance decreases a lot by NNShot, which we attribute to the likely underrepresented representation by dimension reduced operation. However, due to the memory limits, we leave more investigation for further research. Besides, how to efficiently adapt NNShot to long documents is also an open challenge.

\paragraph{Overall Analysis}
We attribute the observed results to three main reasons:  First, the document-level argument extraction involves reasoning over a much longer context compared to the sentence-level. Actually, the average number of sentences for DocEE is $30.71$ as pointed out in Table \ref{tab:general-statistics}, which dramatically increases the difficulty of effective encoding over long documents. Even though Longformer can capture attention over long sentences, the ability is still limited. On the other hand, document-level argument extraction faces the new challenge of extremely unbalanced label distribution with more than $95\%$ of label $\texttt{O}$. Compared to its sentence-level counterparts, where the label unbalance already degrades the performance, the document-level sparse distribution of arguments further exacerbates the unbalanced distribution. The majority of $\texttt{O}$ labels make it difficult for the model to learn a good representation of arguments among the representation space. Besides, to make the few-shot learning close to a realistic setting, we follow the long-tail arguments distribution of the original DocEE dataset. This extremely unbalanced setting is a good testbed for validating the model ability due to its similar distribution of many real-world few-shot problems, as also pointed by \cite{sabo2021revisiting}. As the results suggest, the long-tail distribution makes it hard for models to uniformly focus on all labels. Third, due to the high GPU memory and computation requirements brought by long documents, we only aim at providing benchmark baselines results. More advanced methods might help, but we leave it for future work.

\paragraph{Error Analysis}
Finally, we report the false positive (FP) and false negative (FN) scores in Table~\ref{FN} for two domain using ProtoNet-Longformer. The overall FP results demonstrate that most $\texttt{O}$ are correctly predicted. However, considering the large number of $\texttt{O}$ labels compared with real arguments, even a small portion of FP still leads to a large performance drop of final results.
As for real arguments prediction, the main misclassification errors come from assigning one argument type to another type considering FN is low.

\section{Conclusion}
In order to handle new emerging event arguments with limited annotations and adapt it to the real-world document-level scenario, we propose \methodname~ benchmark to advance the research of few-shot learning for document-level event argument extraction. 
We conduct comprehensive experiments by extending previous models into our task under in-domain and cross-domain. Our experiments confirm the necessity of moving to document level by showing that current models still witnesses suboptimal performance. We also demonstrate the benefits of using a more diverse training base to learn a good argument feature extractor.
The current results show that \methodname~ is challenging due to the long document and limited examples, as well as the intrinsic charisma of few-shot learning. The relatively low extraction score illustrates the difficulty of this novel task, in the meanwhile it also provides new chances for advancing this field. In summary, we hope \methodname~ shed new light on a more realistic but challenging setting for event argument extraction. In the future, we hope to investigate more advanced methods for solving this problem. 

\section{Limitations}
As we mentioned, we only focus on event arguments with the assumption that event type is already provided. However, this is not always true for many applications in real life scenarios. But it would be out of the scope of this work to combine them together, so we leave it for future work. 
Besides, considering the long input of document-level extraction, the computing memory consumption significantly increase to tens of times compared with its sentence-level counterpart. We only consider the 1/2-Doc cases, although in reality more docs are possible. We believe finding a solution for decreasing the memory requirements would be of great impact for future research in this direction.

\section{Acknowledgments}
Xianjun Yang was supported by the UC Santa Barbara NSF Quantum Foundry funded via the Q-AMASEi program under NSF award DMR1906325.

\bibliography{anthology,custom}
\bibliographystyle{acl_natbib}

\appendix

\section{Experiments}\label{experiments}

\subsection{Evaluation Metrics} 
Since we treat this event arguments extraction as a sequence labeling task, we employ $\texttt{IO}$ notation, where all tokens within an argument type are labeled as $\texttt{I-type}$ while all other tokens are labeled as $\texttt{O}$. Besides, we report all the performance on non $\texttt{O}$ types. The prediction is only considered as correct when all tokens within that argument are correctly classified. We use macro precision, recall, and F1 score to measure the performance.

\subsection{Experimental Configuration} 
For all transformer-based models, we employ the released model from HuggingFace\footnote{https://huggingface.co/models} and set the learning rate to $[1e-4, 1e-5, 5e-5]$, of which $1e-5$ is the best parameter except for $3$w$1$d. For 3w1d tasks, we use a learning rate of $1e-6$, otherwise, it will not converge. We try the different batch size of $[1, 2, 4, 6]$, of which 4 and 6 does not lead to converging, $2$ achieves the best performance. We use AdamW \cite{DBLP:conf/iclr/LoshchilovH19} as the optimizer and gradient clipping of $1.0$. We use the NLTK\footnote{https://www.nltk.org/} package for sentence tokenization.
All models can be put into four NVIDIA A$60$ GPUs with an RAM of $48$GB each. The training procedure takes around $15$ hours and $10$ hours for $60$k iterations to complete for $2$-Doc and $1$-Doc settings, respectively. We report the mean and variance for all experiments under two random seeds, except only one run for ProtoNet-MNAV. The validation is done by every $4$k interaction on the training episodes, and we use the best checkpoint from the validation results for testing. The number of query instances during testing is always set to $1$. For ProtoNet-MNAV, we tried the hyperparameter K ranging from 2 to 6 and did not observe obvious difference. We report the results based on $K$=$6$.

\section{Memory and Computation Issues}
Due to the length distribution of this document-level task, we set at least 1,024 tokens as chunk length for input to LED-based models. However, a 2-Doc setting results in 4-doc documents coming from both the query and support sets. Considering the argument-extraction is conducted at every token level, the similarity score calculation imposes a severe memory issues for adapting more complicated methodologies. We believe that this is a big challenge for few-shot document-level tasks, which is not only a issue for small language models \cite{sabo2021revisiting} but also true for large models like GPT-3 \cite{brown2020language} and leave more exploration for future work.

\section{Case-study}\label{case}
\subsection{Predictions vs. True}
\begin{figure*}
\centering
    \includegraphics[width=1.0\textwidth]{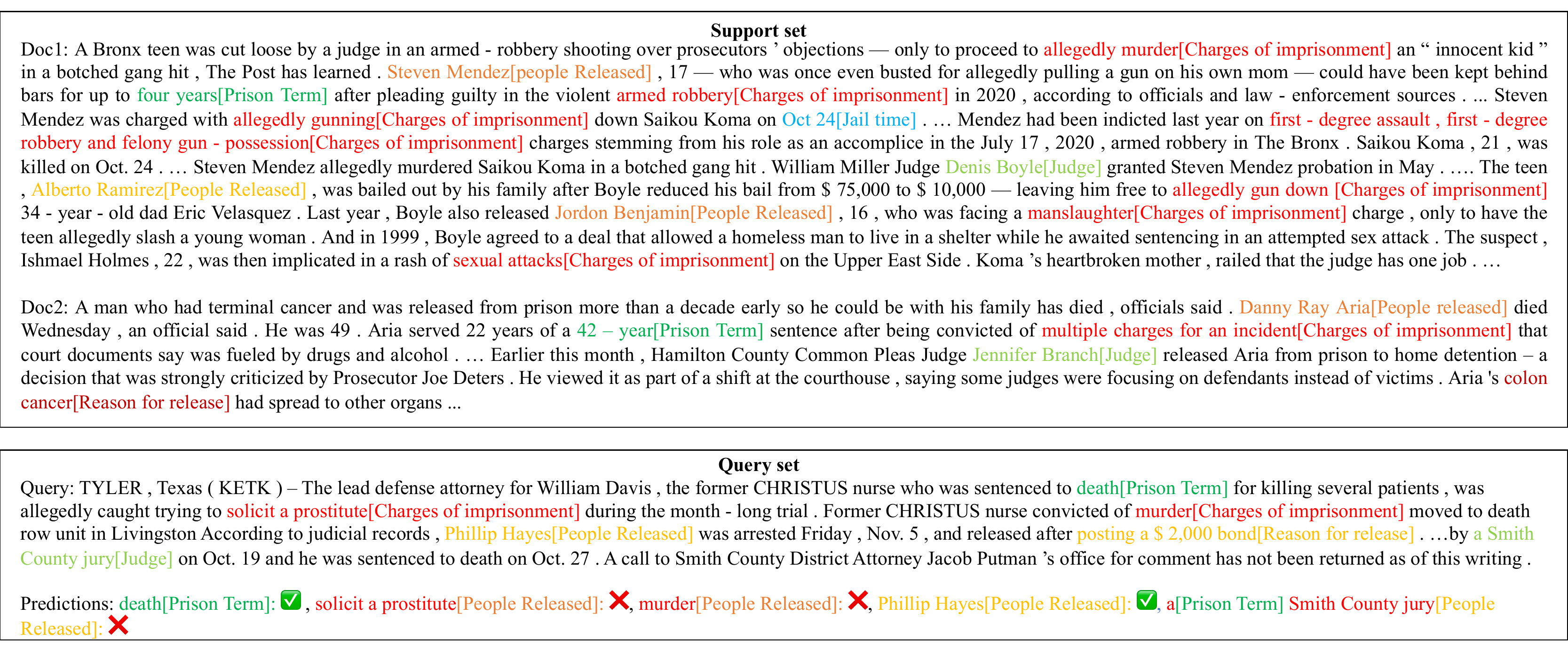} 
    \caption{\label{fig:case1} A case study of a $6$-Way-$2$-Doc episode consisting of a support and query set. For simplicity, we only show part of the documents hereinafter. Best viewed in color.}
\end{figure*}

\begin{figure*}
\centering
    \includegraphics[width=1.0\textwidth]{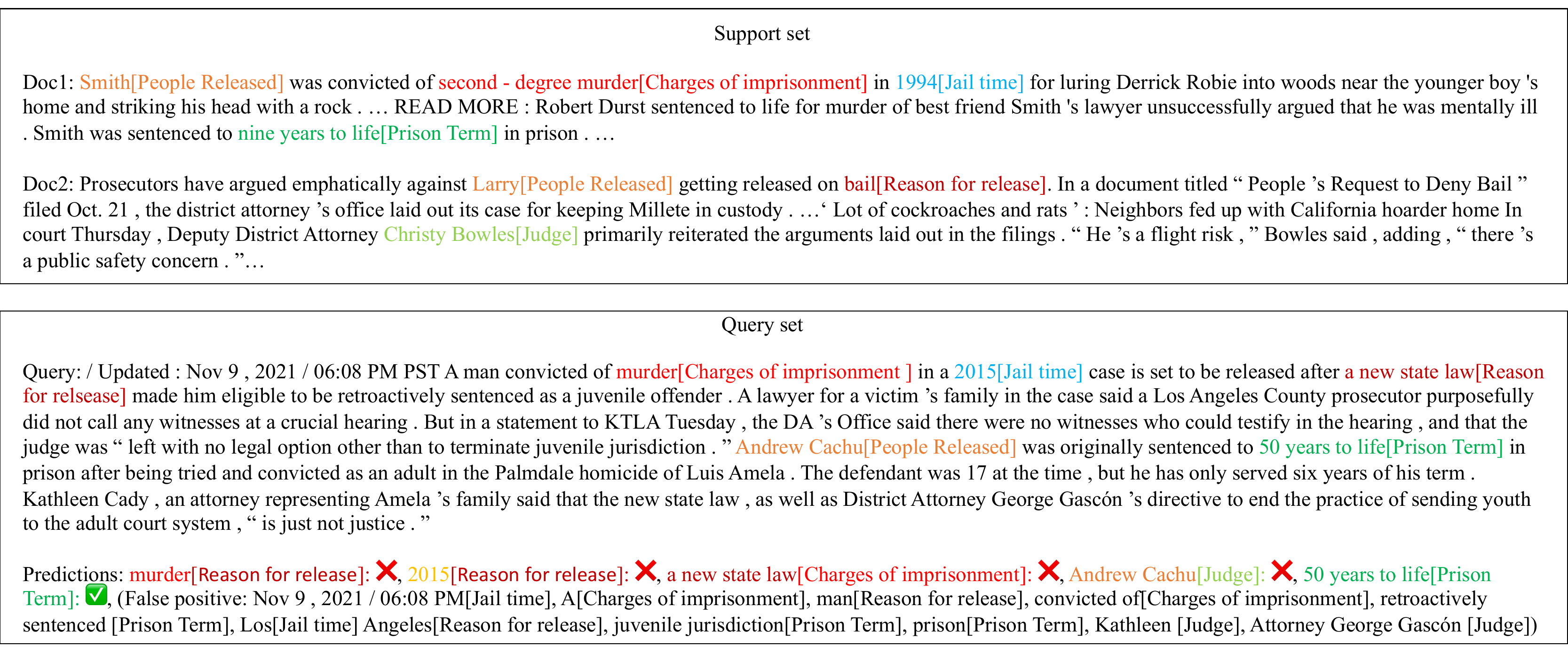} 
    \caption{\label{fig:case2} Another case study of a $6$-Way-$2$-Doc episode consisting of a support and query set. Best viewed in color.}
\end{figure*}

Here in Figure \ref{fig:case1} and Figure \ref{fig:case2} we show two example case study of how our predictions differ from the true labels. The different color corresponds to different event argument types. And we also highlight whether the predictions are accurate or not.
All examples are drawn from the test set in in-domain(samll) setting and the predictions are always the ProtoNet-LongFormer model. As we can see, the majority of predictions are wrong with only a few exceptions in Fig. \ref{fig:case1}. Besides, in Fig. 
\ref{fig:case2} we additionally show the false positive results which also accounts a large protion for the final performance. In general, the current model can not well handle the document-level predictions under few-shot setting, and the prototypical representation for different labels still struggle with token classification.

\subsection{Visualizations}\label{visualizations}
\begin{figure*}
\centering
    \includegraphics[width=1.0\textwidth]{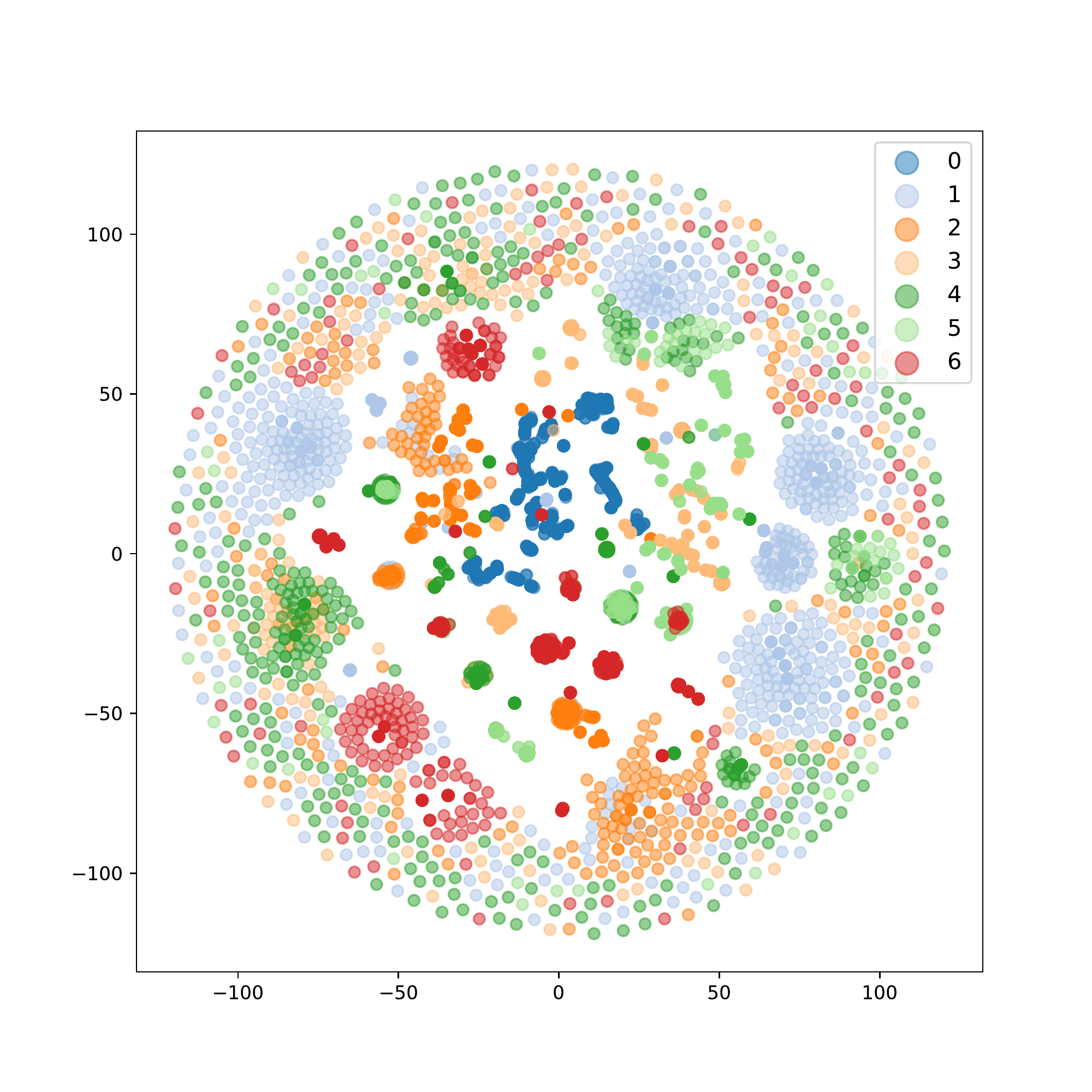} 
    \caption{\label{fig:tsne1} Visualizing prototypical feature vectors using t-SNE. 1 NONA labels(0) and 6 argument types (1 to 5) from trained in-domain(small) checkpoints. Best viewed in color.}
\end{figure*}

\begin{figure*}
\centering
    \includegraphics[width=1.0\textwidth]{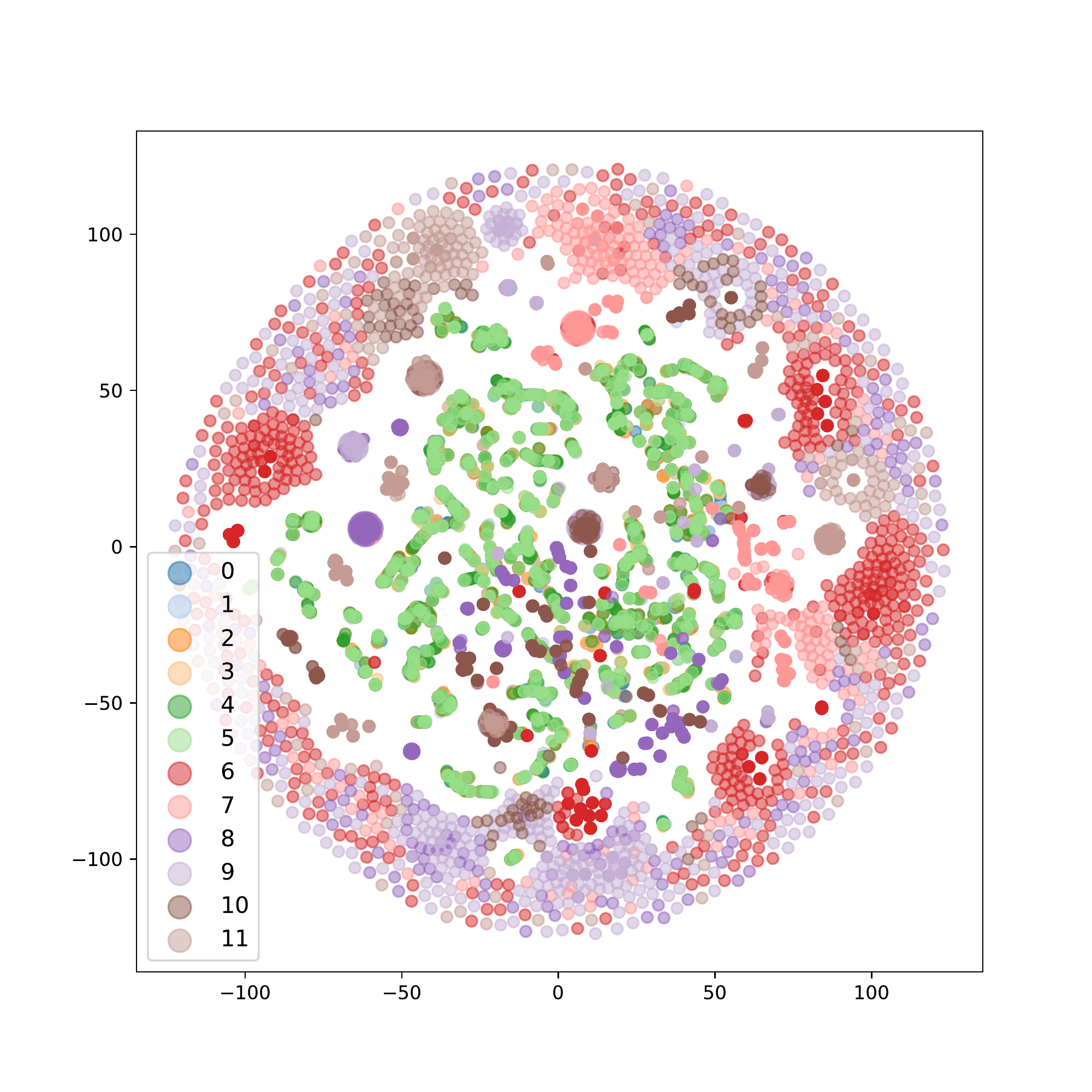} 
    \caption{\label{fig:tsne} Visualizing prototypical feature vectors using t-SNE. 6 NONA labels (0 to 5) and 6 argument types (6 to 12) from trained in-domain(small) checkpoints. Best viewed in color.}
\end{figure*}

Here we plot the two dimensional t-SNE\footnote{https://scikit-learn.org/stable/index.html} projection of the prototypical embeddings in Fig. \ref{fig:tsne} and Fig. \ref{fig:tsne1}. Ideally 7 or 12 clusters as expected for ProtoNet-LongFormer and ProtoNet-MNAV models. But the results show that the clusters are not always well seperated from each other, which might explains the reason why we still get low performance. Actually, we can also clearly see some clusters indeed include many same argument types, like class $0$, $1$ and $6$ in Fig. \ref{fig:tsne1}, but these clusters still spread across the many different locations. For example, we can clearly see 3 big clusters for $6$ and 5 big clusters for $1$, which indicts that multiple prototypes exist for different argument types.
However, when we extend to multiple NONA vectors by ProtoNet-MNAV model, we even see a little performance drop as mentioned in Section \ref{analysis}. Ideally multile NONA vectors can better represent more diverse NONA class but as we can observe from Fig. \ref{fig:tsne} that the multiple NONA vectors(0 to 5) actually make the overall cluster boundary more obscure. The resulting clusters become even more difficult to be classified. 

In general, the current model succeed performing classification for part of the arguments but still fails generating well represented representations for some difficult cases and thus leads to suboptimal resutls. 

\end{document}